# Satellite Image Classification with Deep Learning

Mark Pritt
Lockheed Martin Space
Gaithersburg, Maryland
mark.pritt@lmco.com

Gary Chern
Lockheed Martin Space
Palo Alto, California
gary.chern@lmco.com

*Abstract*—Satellite imagery is important for many applications including disaster response, law enforcement, and environmental monitoring. These applications require the manual identification of objects and facilities in the imagery. Because the geographic expanses to be covered are great and the analysts available to conduct the searches are few, automation is required. Yet traditional object detection and classification algorithms are too inaccurate and unreliable to solve the problem. Deep learning is a family of machine learning algorithms that have shown promise for the automation of such tasks. It has achieved success in image understanding by means of convolutional neural networks. In this paper we apply them to the problem of object and facility recognition in high-resolution, multi-spectral satellite imagery. We describe a deep learning system for classifying objects and facilities from the IARPA Functional Map of the World (fMoW) dataset into 63 different classes. The system consists of an ensemble of convolutional neural networks and additional neural networks that integrate satellite metadata with image features. It is implemented in Python using the Keras and TensorFlow deep learning libraries and runs on a Linux server with an NVIDIA Titan X graphics card. At the time of writing the system is in 2nd place in the fMoW TopCoder competition. Its total accuracy is 83%, the $F_1$ score is 0.797, and it classifies 15 of the classes with accuracies of 95% or better.

## I. INTRODUCTION

Deep learning is a class of machine learning models that represent data at different levels of abstraction by means of multiple processing layers [1]. It has achieved astonishing success in object detection and classification by combining large neural network models, called convolutional neural networks (CNNs), with powerful graphical processing units (GPUs). Since 2012, CNN-based algorithms have dominated the annual ImageNet Large Scale Visual Recognition Challenge for detecting and classifying objects in photographs [2]. This success has caused a revolution in image understanding, and the major technology companies, including Google, Microsoft and Facebook, have already deployed CNN-based products and services [1].

A CNN consists of a series of processing layers as shown in Fig. 1. Each layer is a family of convolution filters that detect image features. In the early layers, the feature detectors look like the Gabor-like and color blob filters shown in Fig. 2, and successive layers form higher-level feature detectors. Near the end of the series, the CNN combines the detector outputs in fully connected "dense" layers, finally producing a set of predicted probabilities, one for each class. Unlike older methods like SIFT [3] and HOG [4], CNNs do not require the algorithm designer to engineer feature detectors. The network itself learns which features to detect, and how to detect them, as it trains.

The earliest successful CNNs consisted of less than 10 layers and were designed for handwritten zip code recognition. LeNet had 5 layers [5] and AlexNet had 8 layers [6]. Since then, the trend has been toward more complexity. In 2015 VGG appeared with 16 layers [7] followed by Google's Inception with 22 layers [8]. Subsequent versions of Inception have even more layers [9], ResNet has 152 layers [10] and DenseNet has 161 layers [11].

Such large CNNs require computational power, which is provided by advanced GPUs. Open source deep learning software libraries such as TensorFlow [12] and Keras [13], along with fast GPUs, have helped fuel continuing advances in deep learning.

## II. PROBLEM

Searching for objects, facilities, and events of interest in satellite imagery is an important problem. Law enforcement agencies seek to detect unlicensed mining operations or illegal fishing vessels, disaster response teams wish to map mud slides or flooding, and financial investors seek better ways to monitor agricultural development or oil well construction. Because the geographic expanses to be covered are great and the analysts available to conduct the searches are few, automation is required. Yet traditional object detection and classification algorithms are too inaccurate and unreliable to solve the problem. What is needed is a deep learning system that can recognize and label objects and facilities automatically, as illustrated in Fig. 3.

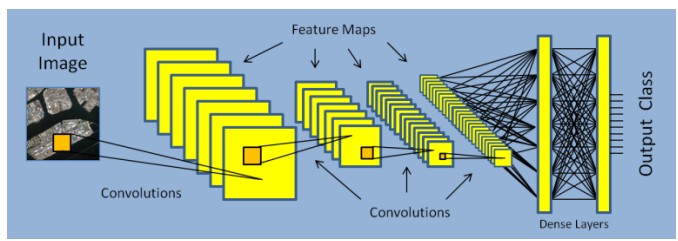

Fig. 1. The structure of a convolutional neural network (CNN). The input image is passed through a series of image feature detectors.

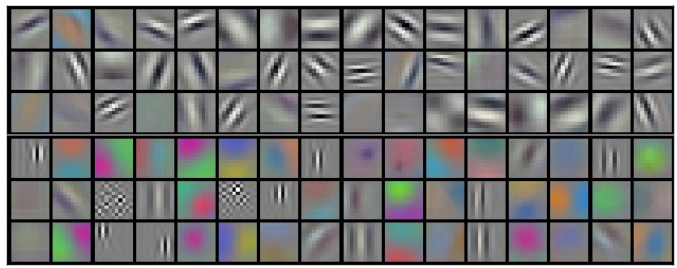

Fig. 2. Examples of the image feature detectors that a CNN might "learn" during its training [6].

Despite a number of recent efforts, deep learning has enjoyed only limited success when applied to satellite imagery. One reason is the difficulty of preprocessing satellite imagery effectively for input to CNNs.

To keep processing times reasonable, CNNs require relatively small fixed-size images. For example, ResNet [10] and DenseNet [11] work with 224x224-pixel images, while Inception [8,9] accepts images of size 299x299. The standard practice in deep learning is to crop and warp the images to the required size [14]. For ordinary photographs, these operations preserve important image features. Such is not the case for satellite images, however, because objects and facilities can be much larger than objects in ordinary photographs. Airports and shipyards, for example, can be tens of thousands of pixels in size. When such large images are resized to a much smaller size of 224x224 or 299x299, small details are lost. These details might include important distinguishing features such as airplanes on a tarmac or container cranes in a shipyard. In addition, there is no way to crop such an image to the smaller size in an effort to preserve small features without losing much of the image. Other complications that plague satellite images as opposed to ordinary photographs include the extremely large aspect ratios of objects such as runways and chicken barns, the wider range of object orientations such as upside-down buildings, and obscuration by clouds as shown in Fig. 4. These conditions make it a challenge to prepare satellite imagery properly for input to CNNs.

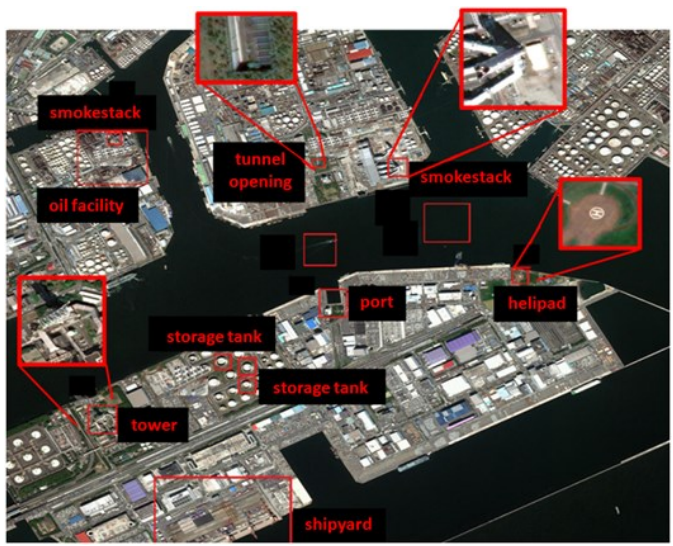

Fig. 3. The results of object and facility recognition with deep learning applied to satellite imagery. The unlabeled boxes are false detections.

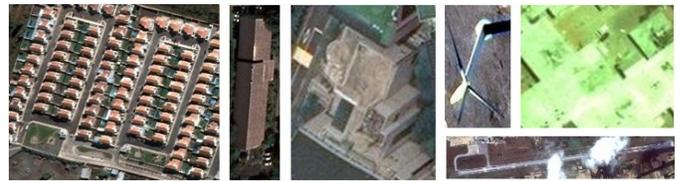

Fig. 4. Objects in satellite images are subject to more extreme viewing conditions than objects in ordinary photographs. They can be very large or very small, long and narrow, upside down, or obscured by clouds.

Perhaps the main reason for the limited success of deep learning on satellite imagery is the lack of large datasets of labeled imagery for algorithm training. Deep learning algorithms typically require thousands of labeled images per class before they are ready to recognize objects in images. Until recently, no such datasets existed.

III. PRIOR WORK

Several annotated datasets of imagery, along with detection and classification activities, have recently appeared. Most of the deep learning applied to remotely sensed imagery has dealt with land cover classification or building detection. For example, the UC Merced Land Use Dataset comprises 2100 aerial images from the U.S. Geological Survey [15,16]. The images are 256x256 pixels in size with a ground sample distance of 0.3 meters per pixel. The 21 classes include land cover types such as agriculture, road, and water, and facility classes such as storage tanks and tennis courts. Several researchers used the VGG, ResNet, and Inception CNNs to classify the UC Merced images into land cover types [17-19] and one reported classification accuracies as high as 98.5% [19]. This dataset is very limited, however, in its size, the number and types of classes, and geographic diversity.

The SpaceNet dataset [20] consists of high-resolution DigitalGlobe satellite images of five cities along with building footprints. CNNs have been trained to segment the images and extract building footprints [21]. This dataset is limited both in its geographic coverage and utility for training a classifier.

Ref. [22] contains a list of other remote sensing datasets. None of them contain the hundreds of thousands of images on a global scale that are required to develop a versatile image classification system.

IV. DATASET

In August 2017 the Intelligence Advanced Research Projects Agency (IARPA) released a very large dataset of labeled satellite imagery called the Functional Map of the World or fMoW [23]. Consisting of one million satellite images labeled in 62 classes, this dataset accompanies a TopCoder challenge [24] that runs from August to December 2017. The challenge is to classify objects and facilities from the satellite images into 62 classes as diagrammed in Fig. 5, plus a separate class of false detections. These classes include airstrips, oil and gas facilities, surface mines, tunnel openings, shipyards, ponds, and towers. The images to be classified include single images as well as sequences of images that comprise an event such as road flooding, construction activity, or debris deposition. Allowances must be made for false detections, since the detection component of an operational object recognition system could produce many false detections.

The fMoW images are available from a publicly accessible bucket on Amazon S3 [25]. They are high resolution, multi-spectral images from the DigitalGlobe constellation of satellites with a nominal ground sample distance (GSD) of 0.5 meter per pixel and four or more spectral bands, including near infrared (NIR). They are accompanied by satellite image metadata, including GSD, date time, bounding box measurements (which define sub-images that contain objects or facilities), acquisition angles, sun angles, and band wavelengths. Fig. 6 shows examples of the images, and Figs. 3 and 4 show examples of the objects and facilities.

The size and global scope of the fMoW dataset are unprecedented, making it ideal for designing deep learning classification systems that work with satellite imagery.

## V. METHODS

We developed the deep learning system of Fig. 7 to classify objects and facilities from the fMoW dataset. The system inputs a satellite image along with metadata, which includes a bounding box that defines the location of an object. It classifies this input into one of 63 classes, including the false detection class.

The system consists of an ensemble of CNNs along with image preparation operations and neural networks (NNs) that combine the image features from the CNNs with the image metadata. The ensemble merges the outputs of the NNs by means of unweighted averaging into a set of prediction probabilities for the 63 classes. The maximum probability determines the classification.

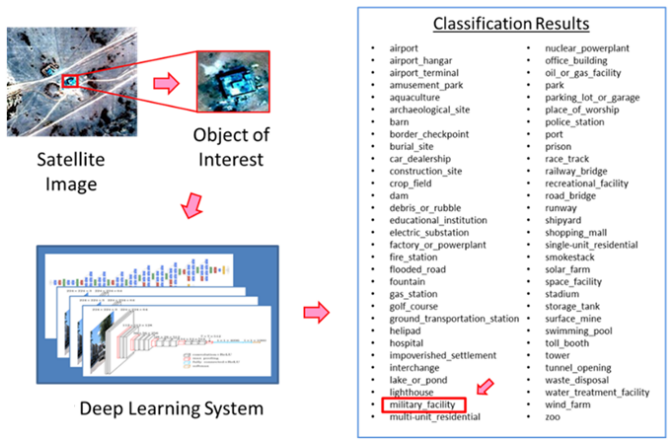

Fig. 5. The IARPA Functional Map of the World (fMoW) challenge is to build a deep learning system, such as the one diagrammed above, that classifies satellite imagery into 62 object and facility classes.

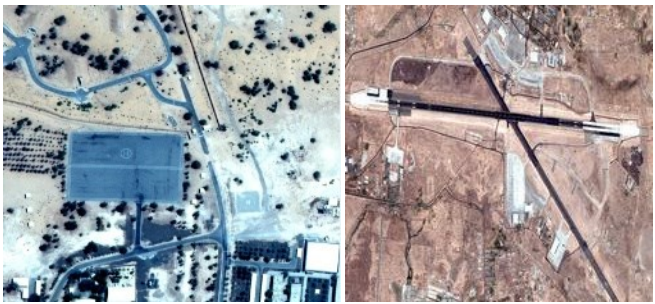

Fig. 6. Examples of the satellite images from the IARPA fMoW dataset.

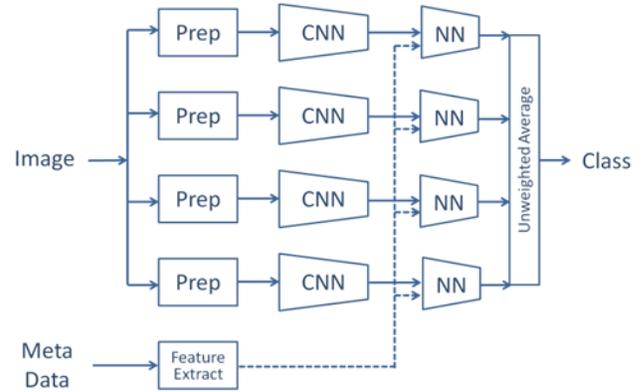

Fig. 7. The deep learning system for classifying satellite imagery.

We implemented the system in Python using the Keras and TensorFlow deep learning libraries. We performed the training and testing on servers running CentOS Linux with NVIDIA Titan X GPUs. After describing each component of the system in detail, we will describe the procedures we followed for training the system.

*A. Image Preparation*

The image preparation step is shown in Fig. 8. The deep learning system, in both its training and operational modes, begins with the input of a satellite image and a bounding box that defines the object or facility to be classified. The image preparation step enlarges the bounding box in proportion to its size, as in [26], before cropping and resizing. This step is necessary to provide the context pixels around the object of interest to the CNN. Alternatively, it squares the bounding box to preserve the aspect ratio of the image features by expanding the smaller dimension to match the larger dimension. (We found that some of the CNNs produced better results with this squaring operation.) It then crops the image to the bounding box and resizes it for input to the CNNs.

*B. CNNs*

After image preparation, the resized images enter the CNNs. Rather than using a single CNN, we use an ensemble in our deep learning system. Each CNN is different in the way it classifies the inputs into classes, and by combining the outputs, significant improvement can be realized. The CNNs consist of DenseNet-161 [11], ResNet-152 [10], Inception-v3 [9] and Xception [27]. At first we used shallower CNNs, including VGG [7] and ResNet-50 [10], but we found that the deeper CNNs produced better results.

*C. Metadata*

The outputs of the CNNs consist of prediction probabilities, one for each of the 63 classes. These probabilities comprise the image features. They are combined with normalized metadata features before they enter the NNs of Fig. 7.

The "Feature Extract" step of Fig. 7 parses the satellite metadata into 27 floating-point values and normalizes them into the range -1 to 1. Table I lists these values along with the computations applied to the metadata before normalization. The purpose of including metadata is to improve the accuracy of classification. Some of the features listed in Table I make sense for this purpose. For example, if the cloud cover percentage is high, there should perhaps be a bias to the false detection class. Others do not seem to make sense. For example, why should the number of bounding boxes influence the classification decision? Our philosophy in selecting these features is to let the NNs decide which features are important.

The NNs have the structure shown in Fig. 9. They combine the 27 metadata features with the 63 image features. These 90 features pass to a 1024-node fully connected middle layer and then to a 63-node output layer. A dropout of 0.60 is applied to the middle layer, and a softmax operation is applied to the final layer to produce the 63 class prediction probabilities.

We experimented with the size and number of layers and found that this configuration worked best. We also experimented with defining the image features from the CNNs differently. For example, in one experiment we accessed the 2048 values in the next-to-last layer instead of the 63 values in the output layer. While this approach worked well, the use of the 63 predictions as the image features worked even better.

### D. Ensemble Learning

Finally, the outputs of the NNs are combined by means of unweighted averaging: the predictions for each class are averaged, and the class with the maximum probability is chosen as the classification. This simple method of ensembling neural network models has proven to be often as effective as more complex ensemble algorithms [28].

### E. Image Sequences

Some of the fMoW data consists of sequences of images rather than single images. Each sequence represents the same object or facility but at different dates and times. For each image, the acquisition angle, sun angle, lighting conditions, and weather conditions can be different. The task is to classify each sequence into a single class. For three of the classes, the sequences represent activities: road flooding, construction, and deposition of debris. We approached this task by separately classifying each image of a given sequence, averaging the predictions for each class over the images, and then taking the maximum prediction as the classification. This approach harmonized well with the unweighted averaging method we used to ensemble the outputs of the networks.

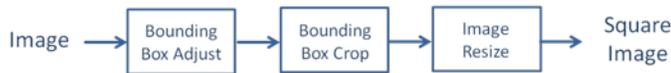

Fig. 8. The image preparation steps of the box labelled "Prep" in Fig. 7.

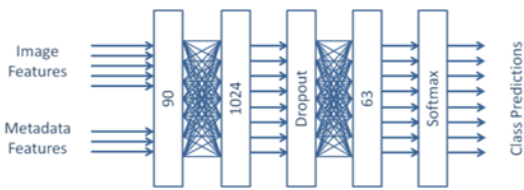

Fig. 9. The structure of the neural networks labeled "NN" in Fig. 7.

TABLE I. METADATA FEATURES EXTRACTED FROM SATELLITE IMAGERY

| Metadata Feature | Computation |
| --- | --- |
| GSD | Ground sample distance in meters |
| Cloud Cover | Percentage |
| Off Nadir Angle | Degrees |
| Longitude (x) | cos(lon) approx. from UTM zone |
| Longitude (y) | sin(lon) approx. from UTM zone |
| Latitude (z) | sin(lat) approx. from UTM zone |
| Year | 2000-2020 |
| Month | 0-11 |
| Day | 0-31 |
| Hour and Minute | 0-23.999 UTC |
| Sun Azimuth Angle (x) | cos(az) |
| Sun Azimuth Angle (y) | sin(az) |
| Sun Elevation Angle | degrees |
| Target Azimuth Angle (x) | cos(az) |
| Target Azimuth Angle (y) | sin(az) |
| Local Hour | Computed from hour and longitude |
| Week Day | Day of the week (0-6) |
| Bounding Boxes | Number of boxes in this image |
| Original Bounding Box Width | Apply base-10 log |
| Original Bounding Box Height | Apply base-10 log |
| Adjusted Bounding Box Width | Apply base-10 log |
| Adjusted Bounding Box Height | Apply base-10 log |
| Bounding Box Aspect Ratio | Apply base-10 log |
| Bounding Box Aspect Ratio | Ratio of min to max, without log |
| Box to Image Width Ratio | Bounding box width / image width |
| Box to Image Height Ratio | Bounding box height / image height |
| Box to Image Ratio | Ratio of min to max dimensions |

### F. False Detection Class

The fMoW dataset included 11,000 false detection images. We split them 90-10% and added them to the training and validation datasets as a 63$^{rd}$ class. We also investigated other methods of defining false detections. In one experiment, whenever the maximum predicted probability failed to exceed a given threshold such as 0.5, we classified the output as a false detection. We found, however, that ad hoc methods like this produced inferior results.

### G. Training

It is common practice in machine learning to split the data into a training set and a validation set. The purpose is to provide data to train the neural networks yet reserve a subset to ensure that they do not over-fit to the training data.

The fMoW data was split into two subsets: a large training subset of 363 thousand images, or 87% of the data, and a smaller validation subset of 53 thousand images. We used the split supplied by IARPA. The 11,000 false detection images were also provided by IARPA but not split into training and validation subsets, so we split them 90-10%. Images with cloud cover greater than 40% or bounding box sizes smaller than 5 pixels, as defined in the metadata, were omitted from the training set.

Our training procedure for the CNNs had several notable features:

- Transfer Learning: Following a common practice called transfer learning, we used CNNs pre-trained on ImageNet images and then fine-tuned them on the fMoW dataset. Some of these pre-trained models are available as part of Keras, while others are provided by the deep learning community [29,30]. The ImageNet dataset consists of photographs and not satellite images, yet CNNs pre-trained on ImageNet have shown the ability to transfer to other image domains [17-19,31,32]. Using pre-trained CNNs as a starting point for fine-tuning on the fMoW images not only reduces the training time from weeks to days but boosts their accuracy.
- Data Augmentation: We increased the size of the training dataset eightfold by means of image flips and 90, 180 and 270 degree rotations as shown in Fig. 10. We found that image shifts and additional rotations of multiples of 45 degrees did not improve the results.
- Epochs: Because of the large amount of data generated by the image augmentation step, we found one epoch to be typically sufficient for convergence.
- Learning Rate: Running a second epoch at a reduced learning rate usually increased the accuracy.

We spent some time trying to train a CNN from scratch, but the fine-tuned networks produced classifications with higher accuracies and were faster to train We also experimented with freezing various layers of the CNNs but did not find significant improvement.

The training of the NNs for integrating the metadata took place after the training of the CNNs and followed a similar procedure. The 63 output softmax probabilities from each CNN and the 27 metadata features were fed into the NNs and trained for 20 epochs with early stopping once the validation loss stopped decreasing. We found that training the NNs separately from the CNNs rather than simultaneously, as others have effectively done [26], resulted in shorter training times and higher accuracies.

## VI. RESULTS AND DISCUSSION

When evaluated on the 53 thousand images of the fMoW validation dataset, the total accuracy of the system is 83% and the $F_1$ score is 0.797. Fig. 11 shows a plot of the precision, recall, and $F_1$ scores of the 63 classes. Fifteen of the classes have accuracies of 95% or better. On the test dataset, the system achieves a TopCoder score of 765,663, which is a weighted $F_1$ score of 0.766. (IARPA weighs certain classes higher or lower than others.) The system is currently in 2[nd] place out of 50 competitors [24].

Fig. 12 shows the confusion matrix of class accuracies. The strong diagonal indicates the power of the system. The off-diagonal elements indicate the classes that it confused with other classes. The strongest confusion involves the shipyard class, which is confused with the port class 56% of the time. In fact, it is difficult even for humans to distinguish shipyards from ports, as shown in Fig. 13. The next pair of classes most likely to be confused is the multi-unit residential class and the single-unit residential class shown in Fig. 14.

The weakest class is office_building with an accuracy of only 24%. Fig. 15 shows examples of this class. It is easily confused with other classes that contain buildings, including fire stations, places of worship, car dealerships, educational institutions, and even barns. Like the case of shipyard vs. port, it can be difficult even for humans to distinguish these classes from one another. The next lowest class is police_station, shown in Fig. 16, with a classification accuracy of 37%.

Certain classes, on the other hand, are easy to identify. The class wind_farm is characterized by very distinctive features as shown in Fig. 17, which enables the system to achieve 99% accuracy. Nuclear power plants, tunnel openings, and golf courses have even higher accuracies. The class crop_field, with its distinctive crop rows and colors as shown in Fig. 18, has a classification accuracy of 97%.

During the fMoW challenge, the Johns Hopkins Applied Physics Laboratory (APL) released their software for image classification [26]. Their solution consisted of the DenseNet-161 CNN with metadata features concatenated with the original global average pooling layer, followed by two additional dense layers and the final softmax layer. The APL solution also featured an LSTM (long short term memory) network for classifying the image sequences. In the fMoW TopCoder competition, this system scored 722,985 points, which corresponded to a weighted $F_1$ score of 0.723. Our system outperformed the APL system by 4.3% with a score of 765,663 points and a weighted $F_1$ score of 0.766. However, their baseline used a single model, not an ensemble. Because their use of an LSTM boosted the accuracy by 1% over the CNN/metadata model, it is possible that an LSTM integrated with our system would show a similar increase in accuracy.

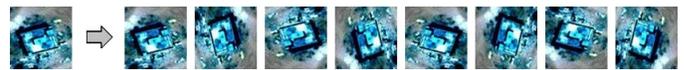

Fig. 10. Data augmentation of the training images by flips and rotations. This increases the effective size of the training dataset by a factor of eight.

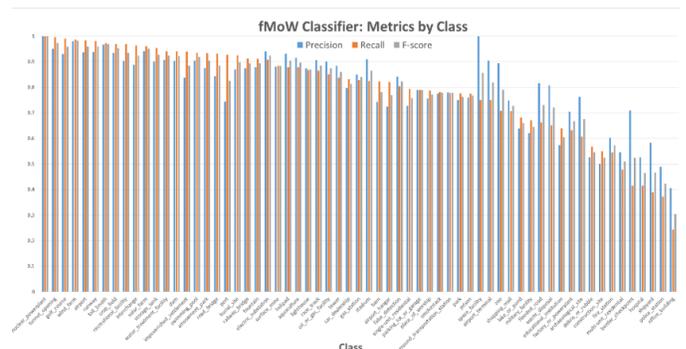

Fig. 11. The precision (blue), recall (red), and $F_1$ score (gray) of the 63 classes. The classes are ordered according to classification accuracy from highest to lowest (left to right).

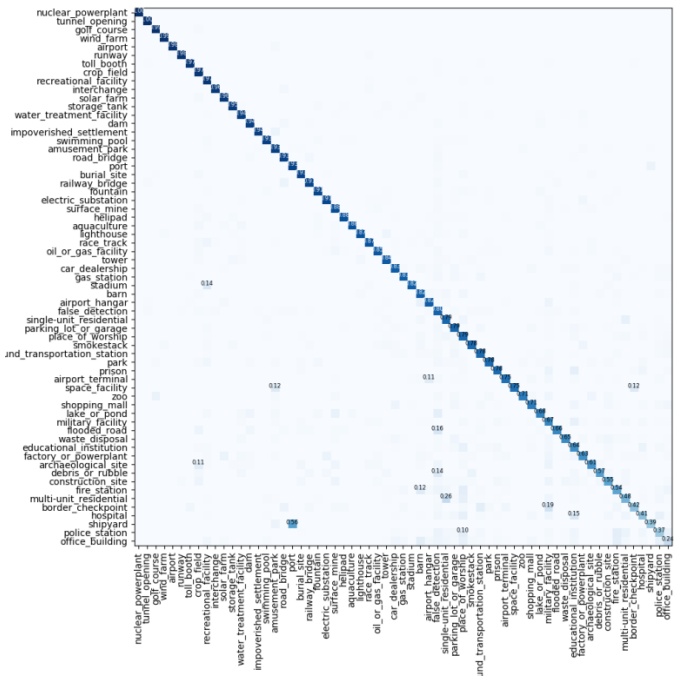

Fig. 12. The 63x63 confusion matrix of the class accuracies. The predicted labels are on the x-axis, and the true labels are on the y-axis. The classes are ordered according to classification accuracy from highest to lowest (top to bottom, left to right).

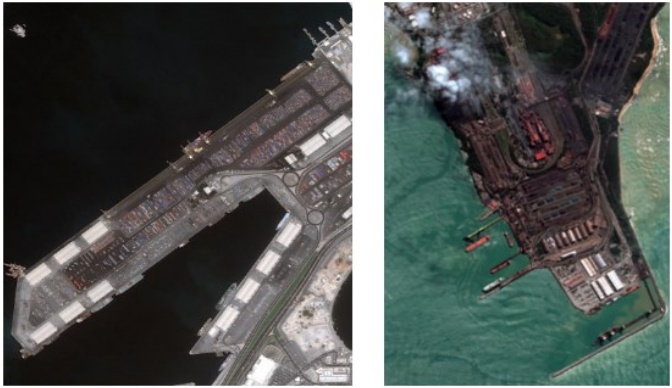

Fig. 13. Shipyards (left) are confused with ports (right) 56% of the time. These two classes are the ones most often confused with one another.

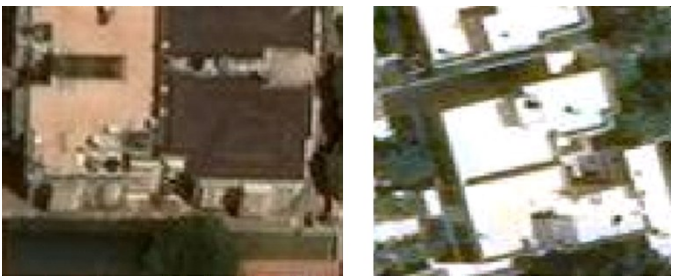

Fig. 14. The multi-unit residential class (left) has a 26% chance of being confused with the single-unit residential class (right).

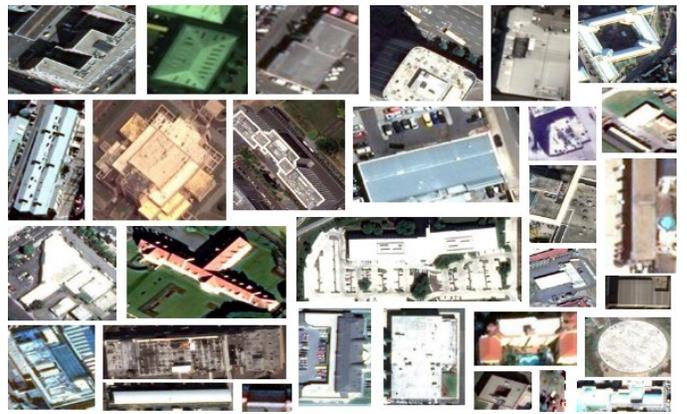

Fig. 15. Examples of the office_building class, which is the class with the lowest accuracy of 24%. Office buildings are easily confused with many other classes, including fire stations, places of worship, car dealerships, police stations (Fig 16) and even barns.

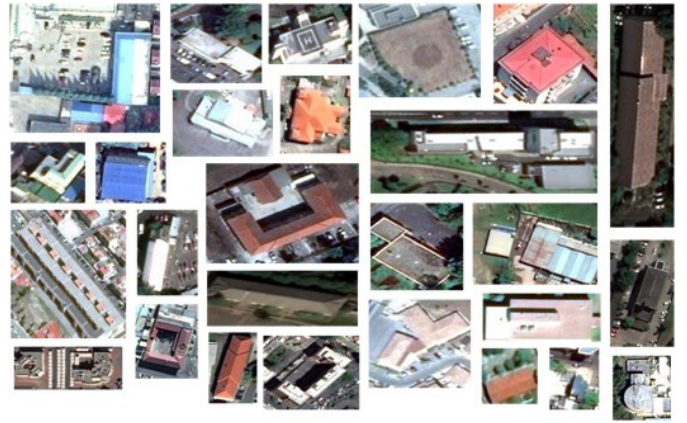

Fig. 16. Examples of the police_station class, which is the class that the deep learning system classifies with the next lowest accuracy of 37%. This class is most often confused with place_of_worship (10%), fire_station (10%) and office_building (7%).

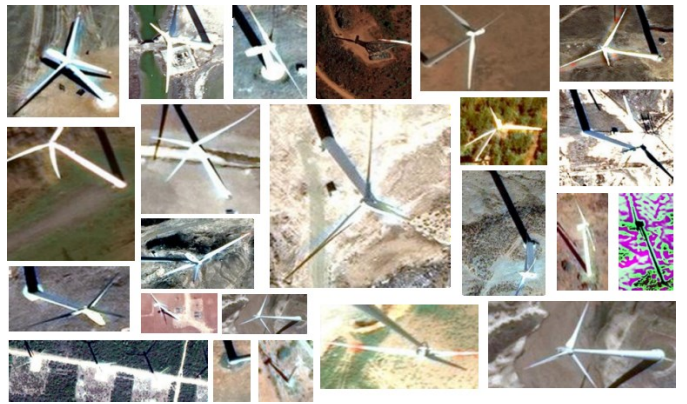

Fig. 17. Examples of the wind_farm class. The distinctive linear features of the towers, fan blades, and shadows enable the deep learning system to classify these images with a very high accuracy of 99%.

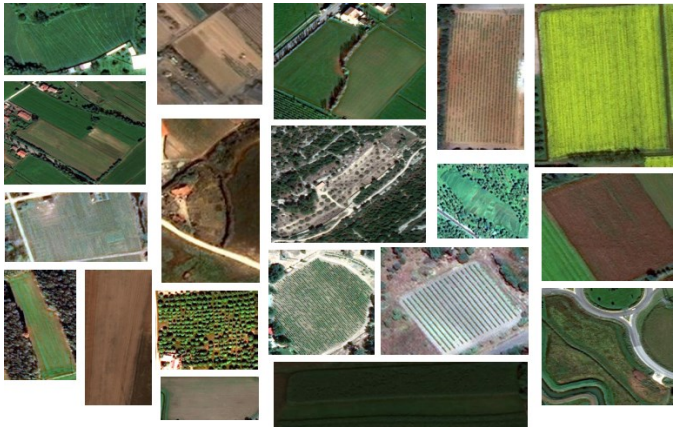

Fig. 18. Examples of the crop field class. This class has an accuracy of 97%.

## VII. CONCLUSION

We have presented a deep learning system that classifies objects and facilities in high-resolution multi-spectral satellite imagery. The system consists of an ensemble of CNNs with post-processing neural networks that combine the predictions from the CNNs with satellite metadata. On the IARPA fMoW dataset of one million images in 63 classes, including the false detection class, the system achieves an accuracy of 0.83 and an $F_1$ score of 0.797. It classifies 15 classes with an accuracy of 95% or better and beats the Johns Hopkins APL model by 4.3% in the fMoW TopCoder challenge.

Combined with a detection component, our system could search large amounts of satellite imagery for objects or facilities of interest. In this way it could solve the problems posed at the beginning of this paper. By monitoring a store of satellite imagery, it could help law enforcement officers detect unlicensed mining operations or illegal fishing vessels, assist natural disaster response teams with the mapping of mud slides or hurricane damage, and enable investors to monitor crop growth or oil well development more effectively.

## ACKNOWLEDGEMENTS

We would like to acknowledge our YellowSubmarine teammates in the fMoW challenge, Brian Gonzalez and Ryan Soldin. We would also like to thank Girard Andres, Mission Systems Capabilities Director at Lockheed Martin, for sponsoring the team.

## REFERENCES

[1] Y. LeCun, Y. Bengio, and G. Hinton, "Deep learning," Nature, vol. 521, pp. 436-444, 28 May 2015.

[2] "Large Scale Visual Recognition Challenge (ILSVRC) – ImageNet," ImageNet, http:// www.image-net.org/challenges/LSVRC.

[3] D.G. Lowe, "Distinctive image features from scale-invariant keypoints," Int. Journal of Computer Vision, vol. 60, no. 2, pp. 91-110, 2004.

[4] D. Navneet and B. Triggs, "Histograms of oriented gradients for human detection," IEEE Computer Society Conference on Computer Vision and Pattern Recognition (CVPR), pp. 886–893, 2005.

[5] Y. LeCun et al., "Backpropagation applied to handwritten zip code recognition," Neural Computation, vol. 1, no. 4, pp. 541-551, 1989.

[6] A. Krizhevsky, I. Sutskever, and G. Hinton, "ImageNet classification with deep convolutional neural networks," NIPS: Neural Info. Proc. Sys., Lake Tahoe, Nevada, 2012.

[7] K. Simonyan and A. Zisserman, "Very deep convolutional networks for large-scale image recognition," arXiv 1409.1556, Sep 2014.

[8] C. Szegedy et al., "Going deeper with convolutions," arXiv 1409.4842, Sep 2014.

[9] C. Szegedy et al., "Rethinking the Inception Architecture for Computer Vision," IEEE Computer Society Conference on Computer Vision and Pattern Recognition (CVPR), 2015.

[10] K. He et al., "Deep residual learning for image recognition," arXiv 1512.03385, Dec 2015.

[11] G. Huang, "Dense connected convolutional neural networks," IEEE Computer Society Conference on Computer Vision and Pattern Recognition (CVPR), 2017.

[12] "TensorFlow: An open-source software library for machin intelligence," TensorFlow, https://www.tensorflow.org/.

[13] F. Chollet at al., "Keras", GitHub, 2017, https://github.com/fchollet/keras.

[14] K. He, X. Zhang, S. Ren, and J. Sun, "Spatial pyramid pooling in deep convolutional networks for visual recognition," IEEE Transactions on Pattern Analysis and Machine Intelligence, vol. 37, no. 9, pp. 1904-1916, Sep 2015.

[15] "UC Merced Land Use Dataset," University of California, Merced, http://weegee.vision.ucmerced.edu/datasets/landuse.html.

[16] Y. Yang and S. Newsam, "Bag-of-visual-words and spatial extensions for land-use classification," Proc. 18th ACM SIGSPATIAL International Symposium on Advances in Geo. Info. Sys., pp. 270-279, 3-5 Nov 2010.

[17] Y. Liang, S. Monteiro, and E. Saber, "Transfer learning for high-resolution aerial image classification," IEEE Workshop Applied Imagery Pattern Recognition (AIPR), Oct 2016.

[18] M. Castelluccio, G. Poggi, and L. Verdoliva, "Land Use Classification in Remote Sensing Images by Convolutional Neural Networks," arXiv 1508.00092, Aug 2015.

[19] G. Scott, M. England, W. Starms, R. Marcum, and C. Davis, "Training deep convolutional neural networks for land–cover classification of high-resolution imagery", IEEE Geoscience and Remote Sensing Letters, vol. 14, no. 4, pp. 549-553, Apr 2017.

[20] "SpaceNet on AWS," Amazon.com, https://aws.amazon.com/public-datasets/spacenet/.

[21] E. Chartock, W. LaRow, and V. Singh, "Extraction of Building Footprints from Satellite Imagery," Stanford University Report, 2017.

[22] G. Cheng, J. Han, and X. Lu, "Remote Sensing Image Scene Classification: Benchmark and State of the Art," Proc. IEEE, vol. 105, no. 10, pp. 1865-1883, October 2017.

[23] "Functional Map of the World Challenge," IARPA, https://www.iarpa.gov/challenges/fmow.html.

[24] "Marathon Match: Functional Map of the World," TopCoder, https://community.topcoder.com/longcontest/?module=ViewProblemStatement&rd=16996&compid=57158.

[25] "Earth on AWS: Functional Map of the World," Amazon.com, https://aws.amazon.com/earth/.

[26] G. Christie. N. Fendley, J. Wilson, and R. Mukherjee, "Functional map of the world," arXiv 1711.07846, 21 Nov 2017.

[27] F. Chollet, "Xception: deep learning with depthwise separable convolutions," arXiv 1610.02357, Oct 2016.

[28] C. Ju, A. Bibaut, and M. van der Laan, "The relative performance of ensemble methods with deep convolutional neural networks for image classification," arXiv 1704.01664, April 2017.

[29] "Applications", Keras, https://keras.io/applications/.

[30] F. Yu, "CNN Finetune", Github, 2017, https://github.com/flyyufelix/cnn_finetune.

[31] A. Razavian, H. Azizpour, J. Sullivan, and S. Carlsson, "CNN features off-the-shelf: an astounding baseline for recognition," IEEE Computer Society Conference on Computer Vision and Pattern Recognition (CVPR), pp. 512-519, 2014.

[32] J. Yosinski, J. Clune, Y. Bengio, and H. Lipson, "How transferable are features in deep neural networks?" Proc. 31st Int. Conf. Machine Learning, vol. 32, pp. 647-655, 2014.